\documentclass[sn-basic]{sn-jnl}%
\usepackage[T1]{fontenc}

\usepackage{paralist}
\usepackage{amsmath}
\usepackage{subcaption}

\newcommand{\object}{\texttt}
\newcommand{\textnl}{\textsl}
\newcommand{\gloss}[1]{`\textnl{#1}'}
\newcommand{\hebgloss}[2]{\gloss{#2}}

\begin{document}

\title[The Knesset Corpus]{The Knesset Corpus: An Annotated Corpus of Hebrew Parliamentary Proceedings}

\author*[1]{\fnm{Gili} \sur{Goldin}}\email{gili.sommer@gmail.com}

\author[2]{\fnm{Nick} \sur{Howell}\email{nlhowell@gmail.com}}

\author[2]{\fnm{Noam} \sur{Ordan}\email{noam.ordan@gmail.com}}

\author[3]{\fnm{Ella} \sur{Rabinovich}\email{ellara@mta.ac.il}}

\author[1]{\fnm{Shuly} \sur{Wintner}}\email{shuly@cs.haifa.ac.il}

\affil[1]{\orgdiv{Department of Computer Science}, \orgname{University of Haifa}, \orgaddress{\country{Israel}}}

\affil[2]{\orgname{IAHLT}, \orgaddress{\country{Israel}}}

\affil[3]{\orgdiv{School of Computer Science}, \orgname{The Academic College of Tel-Aviv Yaffo}, \orgaddress{\country{Israel}}}

\abstract{
We present the Knesset Corpus, a corpus of Hebrew parliamentary proceedings containing over~30 million sentences (over~384 million tokens) from all the (plenary and committee) protocols held in the Israeli parliament between~1998 and~2022. Sentences are annotated with morpho-syntactic information and are associated with detailed meta-information reflecting demographic and political properties of the speakers, based on a large database of parliament members and factions that we compiled.

We discuss the structure and composition of the corpus and the various processing steps we applied to it. To demonstrate the utility of this novel dataset we present two use cases. 
We show that the corpus can be used to examine historical developments in the style of political discussions by showing a reduction in lexical richness in the proceedings over time. 
We also investigate some differences between the styles of men and women speakers. These use cases exemplify the potential of the corpus to shed light on important trends in the Israeli society, supporting research in linguistics, political science, communication, law, etc. 
}


\maketitle

\section{Introduction}
Parliamentary discourse has been the focus of much research in computational linguistics in recent years. 
Corpora that reflect parliamentary proceedings promise immense benefits to research in political science, law, communication, linguistics, and the social sciences in general \citep{Ilie:2017,skubic-fiser-2022-parliamentary}.

Consequently, various efforts around the world focus on the compilation and preparation of parliamentary corpora, in many languages and countries, including 
the multilingual protocols of the European Parliament \citep{koehn-2005-europarl}, 
Canada \citep{beelen2017digitization}, Germany \citep{barbaresi-2018-corpus,BLAETTE18.1024,blaette-etal-2022-germaparl} and the German-speaking world \citep{abrami-etal-2022-german}, Norway \citep{Lapponi2018,solberg-ortiz-2022-norwegian}, Denmark \citep{Danish-parla}, the Czech Republic \citep{hladka-etal-2020-compiling}, Poland \citep{ogrodniczuk-niton-2020-new}, Iceland \citep{runarsson-sigurdsson-2020-parsing}, Slovenia \citep{pancur-erjavec-2020-siparl}, Italy \citep{agnoloni-etal-2022-making}, the UK \citep{NANNI18.6}, and many more.
These efforts benefited from a dedicated workshop, \emph{ParlaClarin}, held three times so far 
\citep{ParlaCLARIN,parlaclarin-2020-parlaclarin,parlaclarin-2022-parlaclarin}.
To the best of our knowledge, ours is the first work that addresses parliamentary corpora in Hebrew.%
\footnote{\href{https://github.com/NLPH/knesset-2004-2005}{A small collection of 282 Knesset protocols} has been previously released \citep{hebrew-lr}.}

We introduce the \emph{Knesset Corpus}, a large dataset of Knesset (Israeli parliament) proceedings. The corpus contains over~30 million sentences, reflecting all the plenary and committee protocols held in the Israeli parliament between the years~1998 and~2022. 
Additionally, we associated the protocols with detailed meta-information reflecting demographic and political properties of the speakers, based on a large database of parliament members and factions that we compiled.

We detail the structure and composition of the corpus, as well as the database of Knesset members, in Section~\ref{sec:corpus}.
The computational processing we employed to construct the corpus is described in Section~\ref{sec:processing}.
Section~\ref{sec:linguistic-annotation} describes the annotation the corpus with linguistic information. 
The annotated corpus is stored in a flexible, easy-to-use database, and we have developed a graphical user interface that facilitates search of and retrieval from the database (Section~\ref{sec:dashboard}).

To demonstrate the utility of this novel dataset for research, especially in the social sciences, we present two use cases (Section~\ref{sec:use-cases}). First, we show that the corpus can be used to examine diachronic developments in the style of political discussions by demonstrating a reduction in lexical richness in the proceedings over time. Second, 
we investigate topical and stylistic differences between male and female speakers, as reflected in the data. We discuss directions for future research in Section~\ref{sec:conclusions}.

The main contribution of this work is thus not only the corpus itself, with its careful organization and associated tools that facilitate human search and exploration, but also the avenues it opens for exploring important trends in the Israeli society, supporting research in various areas of computational social sciences.
The entire corpus, including all annotations and meta-data, as well as the code used to process it, are publicly-available at \href{https://huggingface.co/datasets/HaifaCLGroup/KnessetCorpus}{Hugging Face}.

\section{The Corpus}
\label{sec:corpus}

Our dataset consists of the official proceedings of the Knesset, the Israeli parliament, and includes both plenary sessions (from the years 1992-2022) and committee deliberations (1998-2022). 
Corpus statistics are detailed in Table~\ref{tbl:corpus}.

\begin{table}[hbt]
\centering
\begin{tabular}{lrrr}
 & \multicolumn{1}{c}{\textbf{Plenum}} 
 & \multicolumn{1}{c}{\textbf{Committees}} 
 & \multicolumn{1}{c}{\textbf{Total}} \\
Protocols & 2,817 & 39,246 & 42,063\\
Sentences & 8,026,280 & 24,805,925 & 32,832,205 \\
Tokens & 97,499,854 & 287,090,800 & 384,590,654 \\
Types & 1,140,642 & 1,880,196 & 2,293,065\\
\end{tabular}
\caption{Corpus statistics.}
\label{tbl:corpus}
\end{table}

\subsection{Corpus structure} 
\label{sec:structure}
\begin{figure}[hbt]
\begin{center}
  \includegraphics[width=0.8\textwidth]{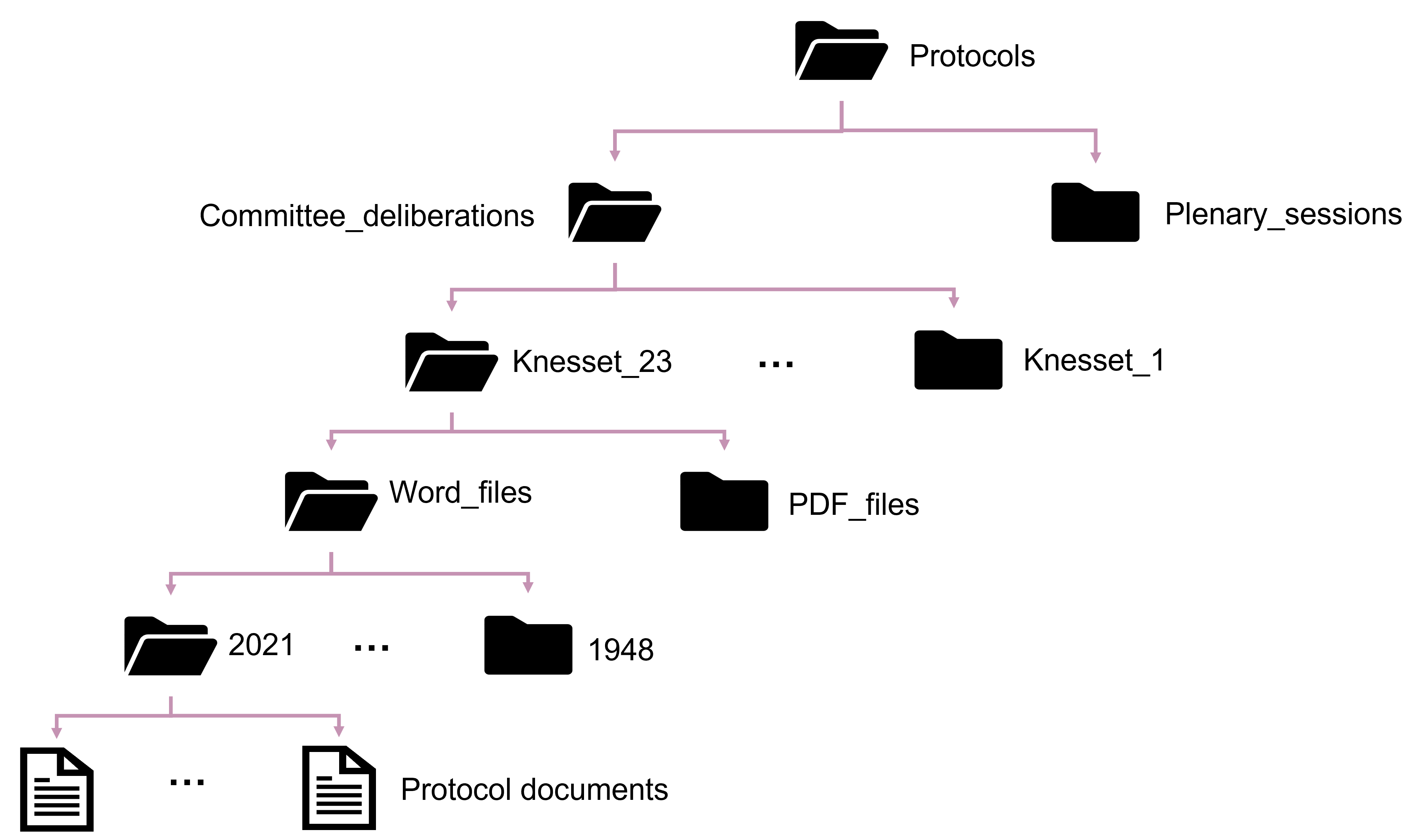}
  \caption{File and folder hierarchy of the corpus.}
  \label{fig:dir_diagram}
  \end{center}
\end{figure}

Knesset proceedings are distributed by the Israeli parliament in the form of Microsoft Word (\texttt{.doc} and \texttt{.docx}) or PDF files, one file per session.
We obtained the files from the Knesset Archives in bulk, and arranged them in organized folders, first by session type (committee or plenary), then by Knesset session,%
\footnote{We use \emph{Knesset session} to refer to the serial number of the parliament, from the 1st Knesset (1948) to the 25th (2022). Each year, the Knesset convenes for two half-yearly conferences, also known as \emph{summer session} and \emph{winter session}. Finally, each meeting of a committee, realized as one protocol in our corpus, is also referred to as a session. This ambiguity should be resolved from the context.}
year, and finally by format (Word files and PDF files). The raw data are released in this structure, see Figure~\ref{fig:dir_diagram}.

We have initially collected all the Knesset documents, from~1948 through~2022, including both plenum and committee protocols. 
The early protocols were 
scanned, OCRed, and converted from PDF to Word (by the Knesset Archives). The later protocols were originally digital (Word files). 
We reviewed a sample of the OCR outputs and manually analyzed their quality. We observed numerous mistakes and inaccuracies, and since the rest of the documents provided a sufficient amount of data, we decided to focus on the originally digital files in this work.  
The remainder of the discussion, therefore, refers only to the committee protocols of the 15th through 24th Knesset sessions spreading over the years 1998-2022, and the
plenary protocols of the 13th through the 24th Knesset sessions, reflecting the years 1992-2022.
For completeness, we distribute also the raw data of earlier proceedings.

\subsection{Meta-data}
\label{sec:meta-data}
In addition to the corpus, we also introduce a database of Knesset members (MPs) and their political affiliations that we have compiled. 
Israel was founded in~1948 and had 25 parliaments since. Each parliament is elected for four years, but can be dissolved earlier. Political parties can relatively easily be categorized according to gross political orientations: left and right orientations often reflect views on security and the Israeli-Arab conflict, rather than social and economic views. Several sectorial parties tend to represent orthodox Jews, ultra-orthodox Jews, Arab citizens, and sometimes other groups. Typically, a coalition of several parties is required in order to form a government.

The MP database contains information about all the Knesset members and all factions represented in the parliament from~1948 to~2022.
This database includes three main tables: one lists all (1100) past and present MPs, with demographic information that includes a unique ID, name, gender, date and country of birth, date of death, 
links to Wikipedia pages, and more.
The second table lists (over~150) factions that were represented in the parliament since its foundation. Each faction is associated with the years (and Knesset sessions) during which it was active, the periods during which it was in coalition, and its political orientation, grossly categorized into eight main classes (extreme left, left, center, right, extreme right; religious; ultra-orthodox; and Arab).
A third table specifies, for each MP, the dates they were affiliated with each of the factions.%
\footnote{Most details about the factions and Knesset members were extracted directly from the Knesset Archives. Specifically, MP  demographic information such as date of birth and death, mother tongue, etc.\ are distributed by the Knesset.
We completed missing details using the \href{https://main.knesset.gov.il/Activity/Info/pages/databases.aspx}{Knesset ODATA website}, the \href{https://oknesset.org/}{Open Knesset project} and \href{https://www.wikipedia.org/}{Wikipedia}. Faction political orientation is based on Wikipedia's 
\href{https://he.wikipedia.org/wiki/קטגוריה:סיעות_בכנסת}{Knesset factions} 
categorization. The meta-data were compiled by a political science graduate student, but the task was quite simple and did not necessarily require an expert.}  

\subsection{Modeling the dataset}
\label{sec:modeling}
We represent the data in JSON format
due to its simplicity, readability, popularity, flexibility and portability.
The dataset consists of four main objects:
\object{Protocol}, \object{Sentence}, \object{Faction}, and \object{Person}.
Each object is associated with rich and extensive information, including both linguistic information (see Section~\ref{sec:linguistic-annotation}) and meta-data (e.g., \object{Protocol}s are associated with their dates, \object{Person}s with their demographic data, and \object{faction}s with their political orientation, as described in Section~\ref{sec:meta-data}).%

Note that this representation is inconsistent with the ParlaMint standards \citep{10.1007/s10579-021-09574-0}. The main difference between our representation of the data and the ParlaMint-TEI standards is the format: ParlaMint-TEI adopts an XML format, organizing data hierarchically from the top level of Parliament down to members and sessions, protocols, sentences and so on, whereas we utilize a JSON format, where each protocol is encapsulated with its sentences and associated data. This choice is motivated by JSON's convenience, offering enhanced flexibility and straightforward access to any data field. However, converting to a ParlaMint-TEI format is only a technical matter which we will consider in future work.

\section{Processing the Data}
\label{sec:processing}
We cleaned the dataset by removing empty or near-empty files, duplicates, and files in which the text was presented graphically.
We then extracted text and meta-data from the files as we now detail.

\subsection{Extracting information from the raw files}
\label{extract_doc}
First, we converted each \texttt{.doc} file to \texttt{.docx} format using Python's \href{https://pypi.org/project/pywin32/}{win32com}
library. We parsed each \texttt{.docx} file as a structured document, using the \href{https://python-docx.readthedocs.io/en/latest/}{docx} package.
Iterating through the documents (and applying adjustments due to the non-uniform nature of the protocols) we extracted the following document-level information:

\begin{compactdesc}
\item[Knesset session] Extracted from the file name. 
\item[Protocol number] Using regular expressions, we looked for relevant keywords such as 
\hebgloss{פרוטוקול מס׳}{protocol no.}
We used the first occurrence since this is usually mentioned as a title opening the document.
\item[Committee name] of committee meetings. We found this in the text by looking for terms such as \hebgloss{מהישיבה}{from the meeting}. 
\item[Protocol date] We located the relevant section in the protocol in which the date was written by searching for keywords such as \hebgloss{יום}{day} and a name of a month. 
We took the first occurrence of such a pattern and used regular expressions to retrieve the date and time components from the paragraph and arrange them in a uniform format.
 
\item[Speakers and their text] We extracted this information using the following algorithm: We assume the first speaker is always the chairperson. To identify the first speaker in the protocol we searched for tokens such as \hebgloss{יו״ר}{chairperson}
which usually come right before the speaker's name. We also made sure the relevant text unit contained a colon and was underlined or otherwise highlighted.

We then matched the speaker's name to the relevant Knesset member \object{Person} object, as given in our meta-data (Section~\ref{sec:meta-data}); if the match failed, we created a new \object{Person} object with this name (see Section~\ref{sec:extract_ppl}).  For each speaker in the protocol, we also indicated whether he or she was chairing the meeting. 
Then, we recorded the current speaker's text and associated it with this speaker.
We continued iterating through the paragraphs of the document, collecting the text of each speaker. We segmented each paragraph into sentences using \href{https://github.com/ufal/udpipe}{udpipe} and if the process has failed for any reason we used \href{https://hebrew-nlp.co.il/service/preprocess/sentencer}{HebrewNLP by Infoneto} instead, making sure each paragraph was indeed segmented to sentences. 
Each sentence was stored as a \object{Sentence} object containing its text and its additional information fields. Each \emph{turn} (a consecutive sequence of sentences by the same speaker) was numbered within the document, and each sentence was numbered within a turn.
\end{compactdesc}

The above conditions were meant to be as general as possible, in order to cover the different ways in which the information appears in the dataset. However, since the documents' structure is highly irregular, we expect that some mistakes may remain in the extracted data. 

\subsection{Processing person information}
\label{sec:extract_ppl}
Identifying the actual Knesset members that occur as speakers in the parliamentary protocols is non-trivial, since they are not associated with IDs in the raw data, and their names may be spelled in a variety of ways. Furthermore, sometimes different Knesset members have identical names.

To link the speaker in the protocol with the correct MP, we used 
regular expressions to ``clean'' the string representing the person's name of extensions such as the person's role or affiliation.
We then applied approximate string matching based on the \href{https://www.drdobbs.com/database/pattern-matching-the-gestalt-approach/184407970?pgno=5}{gestalt pattern matching} algorithm \citep{ratclif:1988}.%
\footnote{We used a modified version of the \texttt{get\_close\_matches} function from the Python \href{https://docs.python.org/3/library/difflib.html}{difflib} module.}
This helped us find the best match for the speaker name, out of our list of MP names,
according to a given threshold. 
We further validated the match by verifying that the person we identified was indeed a Knesset member at the same time as the date of the protocol we were processing.
If so, all the sentences of this speaker in the protocol were associated with the speaker ID of the matched MP. 
If no match was found, either because the speaker name as extracted from the protocol was too different from the name saved in our dataset, or because this speaker was not a Knesset member,
we generated a new \object{Person} object.

\subsection{Evaluation of the extracted data}
We developed the rule-based processing steps outlined above on a small, uniformly-sampled set of 1\% (430 files) of the data. To evaluate the correctness of this process, we defined a small set of~20 randomly selected documents that were not part of the development subset, consisting of~30,636 sentences and~539 speakers. 
We applied the process to this test set and manually checked several artifacts. 
Most of them were correct in 100\% of the test files: the protocol date, first speaker, first sentence and penultimate sentence were all identified correctly. The last sentence was identified correctly in 95\% of the files. Speakers were assigned their correct names in 98.3\% of the cases, and were correctly matched against the MP database in 94.5\% of the cases.
Text that does not indicate speaker names was wrongly identified as a speaker in a negligible number of cases (0.02\%).
We thus trust that the automatically-identified properties of the dataset are highly accurate. The evaluation process and results are detailed in Appendix~\ref{app:evaluation}.

\section{Linguistic Annotations}
\label{sec:linguistic-annotation}

Hebrew, like other Semitic languages, is morphologically rich \citep{semitic-introduction}. Clitics can be both prefixed and suffixed to words. For instance, the definite article and several prepositions are typically prefixed, while pronominal possessive pronouns are often suffixed \citep{,semitic-morphology}. The morphological complexity, along with the writing system, contributes to significant linguistic ambiguity.

With approximately 10 million speakers, Hebrew is relatively well-resourced \citep{hebrew-lr}. Specifically, the availability of linguistically annotated datasets, coupled with robust language models, has been instrumental in enhancing tasks like disambiguation and information extraction \citep{seker-etal-2022-alephbert,zeldes-etal-2022-second,eyal-etal-2023-multilingual,shmidman-etal-2023-pretrained}.
%
%
Table~\ref{tbl:hebrew-corpora} lists some available corpora of Hebrew with their sizes (in millions of tokens), according to \citet{hebrew-lr}. It includes predominantly news articles, plus a very preliminary version of the Knessest corpus with two years of plenary sessions only, unprocessed. To these we add a decade-old dataset of Wikipedia articles, with 3.8M sentences,%
\footnote{\url{https://u.cs.biu.ac.il/~yogo/hebwiki/}.}
and a small corpus of transcribed child- and child-directed language \citep{shai}.
More recently, \citet{Rubinstein2019HistoricalCM} introduced a corpus of Hebrew ``in its formative years'', between~1830 and~1970, consisting mainly of literary texts from the Hebrew equivalent of Project Gutenberg.

\begin{table}[hbt]
\centering
\begin{tabular}{lrl}
 & \multicolumn{1}{c}{\textbf{Tokens (M)}} 
 & \multicolumn{1}{c}{\textbf{Domain}} \\
Haaretz & 11.1 & News \\
Channel 7 & 15.1 & News  \\
TheMarker & 0.7 & News \\
Knesset (old) & 0.2 & Parliamentary \\
Wikipedia & (estimate) 100 & Encyclopedia \\
Childes & 0.4 & Child language \\
Jerusalem Corpus & 22.5 & Literary \\
\hline\textbf{The Knesset corpus} & 384.6 & Parliamentary  \\
\end{tabular}
\caption{Existing Hebrew corpora and their sizes.}
\label{tbl:hebrew-corpora}
\end{table}

In spite of their relative abundance, existing datasets may not adequately address the nuances of a diachronic corpus, particularly in political discourse. 
The sheer size of the Knesset corpus, with its close to 400M tokens (and growing), overshadows existing datasets, and its focus on parliamentary language is unique.

Hebrew treebanks such as HTB \citep{hebrew-treebank, sade-etal-2018-hebrew} and IAHLTwiki \citep{zeldes-etal-2022-second} have significantly contributed to Hebrew linguistic resources, but they are somewhat limited by their respective focus on specific registers and timeframes. The former primarily covers news articles from the late 1990s, providing a limited register scope and a dated linguistic context. Conversely, the latter focuses on contemporary Wikipedia entries, offering a modern perspective yet still within a narrow register.

To address these gaps, we curated a selection of approximately 5,000 sentences from our comprehensive corpus, balancing over various factors, including the type of Knesset protocols (plenary and committee), author gender and political orientation, and the year of publication. We manually annotated this subset with detailed morpho-syntactic information, adhering to the UD V2.10 guidelines \citep{de-marneffe-etal-2021-universal}.%
\footnote{We used the \href{https://arborator.iahlt.org/}{Arborator-Grew annotation tool} \citep{guibon-etal-2020-collaborative}.}
We employed three linguists (two women, one man, all native Hebrew speakers residing in Israel) to annotate the data.%
\footnote{While we did not formally measure inter-annotator agreement, preliminary experiments suggest that these annotated data significantly enhanced model performance; see below.}

For UD parsing we employed a \emph{Trankit}-based model \citep{trankit}. 
Relatively few parsers are available for Hebrew; we chose Trankit because it is a model pre-trained on Hebrew, suitable for the task, which is  known for its exceptional language handling flexibility and its efficiency in performance, especially at inference time, a crucial aspect considering the extensive size of our dataset.\footnote{Our setup included trankit-1.1.1 on Python 3.8 and PyTorch 2.0.1, running on a Tesla T4 (14.75GiB addressable VRAM), completing the task in about 3.5 days. }
Our starting point in training the Trankit parser, and the initial set of trees we used, was Hebpipe \citep{zeldes-etal-2022-second}, significantly expanded with manually-annotated trees from diverse sources (see Appendix~\ref{app:parsing-training}).
We obtained the additional annotated data from the \href{https://www.iahlt.org/}{Israeli Association of Human Language Technologies (IAHLT)};
our unique contribution was the addition of manually-annotated Knesset data. 

We split the extended treebank to 90\% training and 10\% test, stratified by the various data sources. We trained the trankit model on the training portion and tested on the remaining 10\%. 
We used the \href{https://universaldependencies.org/conll18/evaluation.html}{CONLL-2018} evaluation script \citep{zeman-etal-2018-conll}. 
Partial results, listed in Table~\ref{tbl:partial-parsing-results}, indicate very high accuracy.
The full evaluation results are
listed in Appendix~\ref{app:parsing-evaluation} (Table~\ref{tbl:parsing-results}).

\begin{table}[hbt]
\centering
\begin{tabular}{lrrr}
\textbf{Metric} & \textbf{Precision} & \textbf{Recall} & \textbf{F1 Score} \\
UPOS &98.93&98.96&98.94\\
UAS &97.78&97.81&97.79\\
LAS &97.16&97.18&97.17\\
MLAS &91.81&91.86&91.83\\
\end{tabular}
\caption{UD Parser performance.}
\label{tbl:partial-parsing-results}
\end{table}

\begin{figure}[hbt]
\centering
    \includegraphics[width=0.8\textwidth]{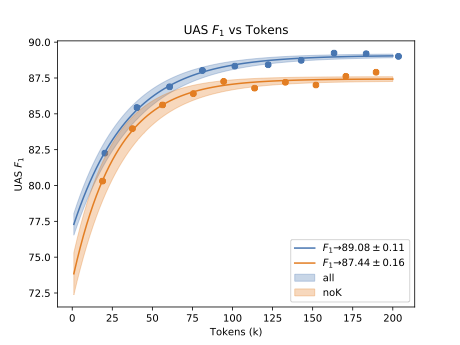}
    \caption{Performance curve for models trained with and without the Knesset data, UAS $F_1$.}
    \label{fig:ud-curve-uas}
\end{figure}

To evaluate the impact of our annotated Knesset data we follow the methodology of \citet{breton}: we plotted learning curves for two models, one (\emph{noK}) excluding Knesset data and the other (\emph{all})
including 90\% of it. Both models were trained on a reduced corpus of~12,682
trees (203k tokens) to optimize computing resources, and both were evaluated on the same set of (10\%) Knesset sentences. We analyzed various UD
evaluation metrics and their learning curves as a function of training corpus size.
Figure~\ref{fig:ud-curve-uas} depicts the $F_1$ performance curve of the UAS (unlabeled attachment score) measure;
other measures, including UPOS (part-of-speech tag), UFeats (morphological features), and MLAS (head plus dependency relation plus features and POS-features-deprel of functional children) are shown in 
Appendix~\ref{app:parsing-evaluation}.

These evaluations underscore the significant contribution of including Knesset data: It yielded substantial improvements across all measures, highlighting the value of domain-specific annotations in enhancing model performance. The observed increase in asymptotic performance reinforces the notion that specialized data can provide benefits unattainable with more generalized datasets.

\section{Corpus Storage and Dissemination}
\label{sec:dashboard}
We chose \href{https://www.elastic.co/}{ElasticSearch} as the database for storing and distributing the Knesset corpus. \href{https://www.knowi.com/blog/what-is-elastic-search/}{ElasticSearch}
is a distributed, open-source search and analytics engine built on Apache Lucene and developed in Java. It facilitates storage, search, and analysis of huge volumes of data quickly and efficiently. 
%
We chose this platform primarily because it supports both flat and nested data structures. This is especially useful for the lexical features of the sentences in our corpus, which would be difficult to represent in a flat tabular format. Another reason is its REST API which makes it easily accessible, along with its GUI platform, \href{https://www.elastic.co/kibana/}{Kibana},
which enables users to explore, query, visualize, and retrieve the data even without technical knowledge. All mentioned features, including linguistic and morphological fields as well as Universal Dependencies (UD) trees, are accessible and can be queried in the database. UD trees are represented as lists of dictionaries, rather than via a graphical interface.

We created a Kibana dashboard for the Knesset corpus that visualizes the data and helps explore it. The dashboard displays information such as the number of sentences stored in the database and their distribution between committee and plenary session, speaker demographics, and more. 
Examples screenshots of the dashboard are shown in Appendix~\ref{app:dashboard}.
The Kibana dashboard and the ElasticSearch database, including their search tools, are publicly available and open, facilitating both human search and code-based querying.

\subsection{The released data}
We release the raw Knesset data, including all \texttt{.doc}, \texttt{.docx} and \texttt{.pdf} documents in our possession, from the 1st Knesset to the 24th, organized as in Figure~\ref{fig:dir_diagram}. 
We also release the processed data, one JSON file per protocol,
in the structure described in Section~\ref{sec:modeling}. 
The original tables containing the meta-data of Knesset members and Knesset factions are available in \texttt{csv} format. 

We release the morpho-syntactic annotations in CONLLU format, along with essential information like sentence and speaker IDs. We also release the manually-annotated set of~5,000 sentences, in the same format.
All resources and code are publicly available at \href{https://huggingface.co/datasets/HaifaCLGroup/KnessetCorpus}{Hugging Face}. and are released under the \href{https://creativecommons.org/licenses/by-sa/4.0/deed.en}{Creative Commons Attribution-ShareAlike 4.0 International License}.
The trained parsing models are licensed freely for research purposes.
In addition, users can access the entire dataset through the ElasticSearch portal, offering programmable querying and downloading via the ElasticSearch REST API, with support for popular programming languages through libraries like the \href{ https://pypi.org/project/elasticsearch/}{ElasticSearch library for Python}. Additionally, the Kibana API enables data visualization, querying, and downloading. Lastly, our Kibana dashboard is also available for exploration. 

\section{Use Cases}
\label{sec:use-cases}
The Knesset Corpus is an invaluable resource for various investigations in political science, law, communication, linguistics, and more. To demonstrate its potential, we describe two simple use cases conducted with this richly-annotated resource.

\subsection{Diachronic changes in linguistic style}
\label{sec:use-case-vocabulary}
Does the style of language used in parliamentary proceedings change over time? Prior studies on diachronic language change in scientific articles \citep{ZHOU2023101262} and literary works \citep{Stajner-12} show a systematic trend of lexical richness and diversity increase over time. 
Here we hypothesize that one possible change would be a deterioration in the richness of the vocabulary used by Knesset speakers, reflecting a general trend of \textit{reduced formality} in such discussions.
Additionally, we expect the language in committee protocols to be less diverse compared to plenary protocols, as the former mainly consists of spontaneous spoken language while the latter mostly consists of prepared, potentially edited speeches.

To investigate these hypotheses we employed two measures: word frequency and type-token ratio (TTR). These measures have been successfully employed for lexical diversity analysis in the context of dementia and aphasia detection \citep{vocabchecker, cunningham2020measuring}, bilingual competency \citep{daller2003lexical, staples2016understanding}, and even male vs.\ female language \citep{litvinova2017differences}. Higher mean word frequency implies higher usage of common words, i.e., lower lexical richness. In contrast, higher TTR implies more unique words out of the total number of word tokens, indicating richer, more diverse vocabulary.
Therefore, we expected (1) higher TTR and lower word frequencies in the plenary data compared with the committee data; and (2) decreasing TTR and increasing word frequencies over the years, especially in the committee sessions.

The two measures---word frequency and TTR---were computed per Knesset annual session (summer and winter sessions of each year).\footnote{More precisely, this is an approximation that relies on fixed months: the ``winter session'' includes data from October to March and the ``summer session''  from April to September.} 
The mean number of protocols collected per session is relatively high%
\footnote{We list the mean with the standard deviation.}
(committee: 798$\pm$375, plenary: 46$\pm$16), as is the average number of distinct speakers (committee: 9534$\pm$4959, plenary: 746$\pm$261). While summer sessions are systematically shorter than winter ones, no other artefacts, potentially confounding the measures, were detected per metadata inspection. Collectively, these findings imply the high robustness of both our metrics.

For each measure, we iterated through the corpus text with a window of~1000 tokens. Within each chunk of~1000 words, we calculated the chunk TTR. We also retrieved the word frequency of each word in the chunk from the \emph{wordfreq} library \citep{robyn_speer_2022_7199437} and averaged the word frequency per chunk.%
\footnote{The \emph{wordfreq} library estimates word frequencies  based on lists that reflect words that occur at least once per 100 million words. The word list for Hebrew was collected from Wikipedia, Open Subtitles, Twitter, and more.}
Finally, we averaged the word frequency and TTR by considering all the chunks in the same Knesset annual session (summer and winter conferences each year) as a unit. We collected the data separately for plenary and committee protocols.

\begin{figure}[hbt]
\centering
\includegraphics[width=\textwidth]{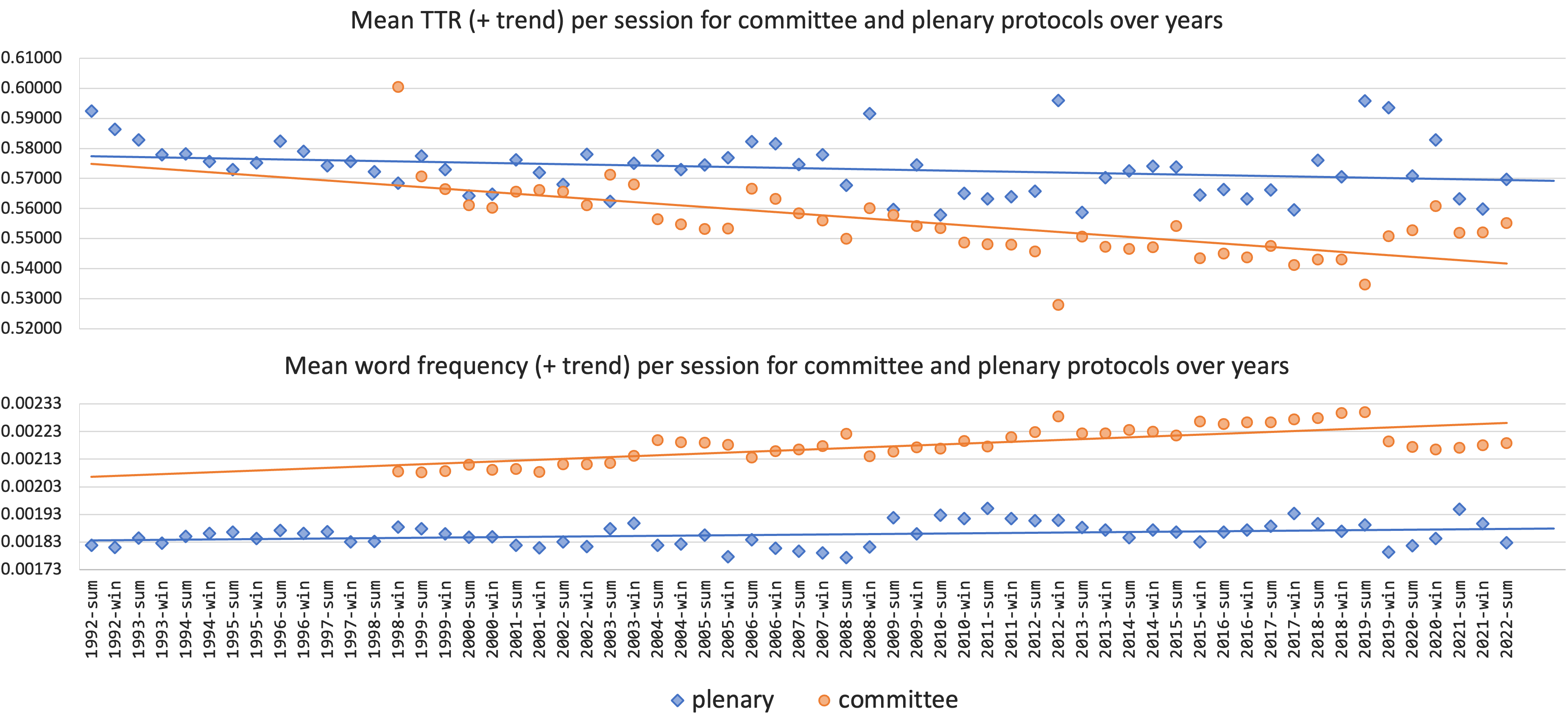}
\caption{Average TTR and word frequency in Knesset plenary and committee sessions. Statistically significant evidence (at $p{<}0.001$) for decreasing TTR was observed for both committee and plenary sessions, and for increasing word frequency in committee protocols. Expectedly, higher TTR and conversely, lower word frequency, were found for the (less formal) plenary sessions. }
\label{fig:ttr}   
\end{figure}

The results of this analysis are graphically depicted in Figure~\ref{fig:ttr}.
The blue markers reflect plenary sessions (1992 to 2022, 61 data points) and the orange markers represent committee meetings (1998 to 2022, 48 data points). 
Each marker corresponds to a Knesset annual session, and the trend line indicates the general direction of the data over time, based on linear regression.
To determine whether there was a significant trend in the results, we conducted the \emph{Mann-Kendall} test \citep{Hussain2019pyMannKendall} at a significance level of 0.05. 


The results confirmed our expectations, revealing statistically significant evidence of decreasing TTR in both plenary ($p{=}0.0029$) and committee ($p{<}10^{-8}$) protocols over the years. Furthermore, the TTR values were lower in the committee protocols compared to the plenary protocols. Additionally, as predicted, the mean word frequency in the committee data exhibited an increasing trend ($p{<}10^{-8}$). However, the trend of increased word frequency in plenary texts was not statistically significant ($p{=}0.0514$).

\subsection{Gender-based differences}
\label{sec:use-case-gender}
Much research indicates that men and women use language differently, to the extent that simple classifiers can distinguish between texts of men and women \citep{Koppel-etal2003,Argamon-etal2003-gender}. \citet{lakoff:1973} pointed out that there was a discrepancy between English as used by men and by women. She indicated differences in vocabulary choices and frequency of lexical items. 
\citet{schmid:2003} compared the  frequencies of words and collocations in text spoken by women and by men; the lexical frequencies that were found were in line with widespread stereotypes about favorite female and male topics.
\citet{Xiufang:2013}  attributed the differences between the language of men and women, \textit{inter alia}, to differences in vocabulary and syntax: women adhered more to grammatical and stylistic conventions.
\citet{muchnik:gender} studied various aspects of gendered language in Hebrew,  but did not provide specific observations regarding the (lexical and syntactic) differences between the Hebrews of men and women, and her research was not based on a large-scale corpus study.

Our second use case leverages two crucial properties of the Knesset corpus: its unique, extensive meta-data (here, information about the gender of Knesset members);%
\footnote{We obviously acknowledge that gender is not binary; we are limited to two genders here because of the binary information on Knesset member genders available to us.}
and the fact that the text is linguistically annotated (specifically, with morphological features such as word lemmas and verb voice).
We focus on gender-based difference in the speech style of Knesset members, and pose two hypotheses.
First (Section~\ref{sec:gender-topcis}), we hypothesize that men and women tend to address different topics in their speeches; and second (Section~\ref{sec:gender-voice}), we hypothesize that they use verb voice differently.%
\footnote{Note that the Knesset corpus is extremely unbalanced: it includes 16,623,263 sentences authored by male MPs, and only 3,720,868 by female MPs. This mirrors real-world gender disparities in political participation and representation.}

\subsubsection{Gender-based topical differences}
\label{sec:gender-topcis}
Differences in the topics addressed by men and women can be indicated by the most prominent words used by authors of one gender but not the other. 
To identify such words we used log odds ratio informative Dirichlet prior \citep{monroe2008fightin}, which has been frequently shown as effective in similar NLP tasks \citep[e.g.,~][]{kessler-2017-scattertext,rabinovich-etal-2019-codeswitch,kumar-etal-2019-topics,kawintiranon-singh-2021-knowledge}. 

We leveraged the availability of morphological annotation in our corpus and focused on lemmas, rather than word forms.%
\footnote{Hebrew is a morphologically-rich language with grammatical gender. Analysing surface forms would run the risk of irrelevant differences that emerge as a result of, e.g., gendered verbs that agree with their subjects.}
Crucially, we leveraged the meta-data that provided the gender of most speakers, and tallied the counts of each lemma in the corpus by gender.
For the log-odds computation we used only lemmas with more than~30 occurrences in each gender class, to eliminate the effect of (unrepresentative) rare words that appear only in one gender class. As a prior for the calculation we used all lemmas in the entire corpus. This process yielded a list of all the lemmas alongside their log-odds values: high values are strongly associated with the language of men, and low,  negative values are more representative of women.%

As this is a qualitative rather than a quantitative exploration, we note that several words stand out immediately as indicating different topics addressed by both classes of authors.
The full lists are available in Appendix~\ref{app:word-lists} (Figure~\ref{fig:word-lists}).
For example, among the top-50 words associated with female authors we found \hebgloss{מגדרי}{gendered}, \hebgloss{מיני}{sexual}, \hebgloss{הטרדה}{abuse}, \hebgloss{אלימות}{violence}, \hebgloss{בריאות}{health}, \hebgloss{זנות}{prostitution}, \hebgloss{רווחה}{welfare}, \hebgloss{מקלט}{shelter}, 
\hebgloss{התמכרות}{addiction},
and more. These all indicate a focus on social, especially woman-related, welfare topics.
Also included are \hebgloss{הורה}{parent}, \hebgloss{ילד}{child}, and \hebgloss{לידה}{birth}, all related to parenting.
In contrast, the top-50 words associated with male authors included \hebgloss{רכב}{vehicle}, \hebgloss{מס}{tax}, \hebgloss{בנק}{bank}, \hebgloss{כביש}{road}, and \hebgloss{ויכוח}{argument}, 
indicating topics `traditionally' associated with men; and \hebgloss{ערבי}{Arab}, \hebgloss{פלסטיני}{Palestinian}, \hebgloss{יהודי}{Jewish}, and \hebgloss{שלום}{peace}, 
indicating security-related topics.
Furthermore, we found an abundance of words that are indicative of the procedural aspects of the Knesset work (\hebgloss{חבר}{member}, 
\hebgloss{מפלגה}{party}, \hebgloss{ממשלה}{government}, \hebgloss{סעיף}{article}, \hebgloss{חוק}{bill (law)}, \hebgloss{סיעה}{faction}, \hebgloss{הצבעה}{vote}), perhaps as a result of most Knesset chairpersons being male.

Interestingly, this analysis also shows some stylistic (as opposed to topical) differences. For example, the top list of female words includes \hebgloss{כן}{yes}, whereas the male list includes \hebgloss{איננו}{isn't} and \hebgloss{נגד}{against}. Also, women overuse \hebgloss{מאוד}{very}, \hebgloss{באמת}{really}, \hebgloss{בעצם}{in fact}, and more hedges.
We leave a deeper investigation of these differences to future research.

\subsubsection{Gender-based differences in voice}
\label{sec:gender-voice}
Evidently, men and women also differ in their style. We test the hypothesis that men and women make different use of verbs in active and passive voice. 
\citet{garera-yarowsky-2009-modeling} found that adding a feature reflecting the percentage of passive verbs used in a text improved the accuracy of a (binary) gender classifier. They did not, however, specify if active voice was found more frequently in male or female language in their data.

Taking advantage of the morphological annotation of our dataset, we computed the frequency of verb voices used by men and women in the entire corpus (the values are \emph{active}, \emph{passive}, and \emph{middle}, while some~13\% of the verbs are unspecified for voice). The results are listed in Table~\ref{tbl:voice}.

\begin{table}[hbt]
\centering
\begin{tabular}{lrrrr}
 & \multicolumn{2}{c}{\textbf{Women}} 
 & \multicolumn{2}{c}{\textbf{Men}} \\
Active & 5,747,960 & 75.39\% & 24,331,402 & 74.93\% \\
Passive  & 370,588 & 4.86\% & 1,719,004 & 5.29\% \\
Middle  & 507,129 & 6.65\% & 2,103,190 & 6.48\% \\
Other  & 999,078 & 13.10\% & 4,318,261 & 13.29\% \\
\end{tabular}
\caption{Usage of verb voice by men and women.}
\label{tbl:voice}
\end{table}

Interestingly, our  data show that women use active verbs \emph{more} than men, and passive verbs \emph{less}. \href{https://www.statsmodels.org/stable/generated/statsmodels.stats.proportion.proportions_ztest.html}{Two proportion z-test}
analysis confirmed these differences were statistically significant ($p{<}10^{-8}$). 

To further validate these results, and to ascertain they were not due to confounds such as gender-related topics that call for more active verbs, we extracted the~20 most frequent active and passive verbs in the text of each gender class. 
The top active verbs were very similar in the text of both genders, and included \hebgloss{רצה}{want}, \hebgloss{אמר}{say}, \hebgloss{חשב}{think}, \hebgloss{ידע}{know}, and \hebgloss{עשה}{do}. The passive verbs were also very similar for the two genders, including \hebgloss{מדובר}{said}, \hebgloss{כתוב}{written}, \hebgloss{התקבלה}{received}, \hebgloss{ניתן}{given}, and \hebgloss{נעשה}{done}. This strengthens our conclusion that the observed differences in voice usage between men and women are not topic-dependent, but rather mirror stylistic differences.


\section{Conclusion and Future Work}
\label{sec:conclusions}
We presented a large corpus of Hebrew parliamentary proceedings, cleaned and organized, with a small subset manually annotated morpho-syntactically; a database of Knesset MPs and factions with detailed demographic and political information; a GUI for accessing and querying the data; and models for UD parsing adapted to this corpus.
We also demonstrated how this novel collection of resources can be used for research in  computational social sciences and digital humanities.

Our future plans include multiple directions. We are currently working on manual annotation of a subset of the corpus for factuality, according to a novel, detailed scheme that we are developing. This annotation will facilitate research on pertinent questions in political science and communication. Specifically, we aim to explore the common belief that the style, but also the contents, of parliamentary discussions (mirroring general societal trends) become more extreme over the years. 

To this end, our current work includes exploration of diachronic dynamics along multiple emotional dimensions, as reflected in MPs' language. We will use VAD (valence, arousal, dominance) emotion models to track temporal trends in political discourse over the years; this work involves compiling high-quality emotion lexicons for Hebrew and tuning Hebrew LLMs on the Knesset dataset. Diachronic topic modeling of the parts of the corpus is another promising direction, capturing (major and subtle) topical trends in the Israeli political discourse. Through these analyses, we aim to provide a detailed insight into the various patterns in which the Hebrew parliamentary discourse evolved in the last decades. 

\subsection*{Acknowledgments}
We are grateful to Inda Novominsky, the  Knesset Archives Director, and to Shmuel Kochav, the Knesset Archives Head of Digitisation and Automation, for making the raw protocols available to us. 
We also thank Maya Cohen-Rahamim, the Knesset Head of Protocols Department and Dikla Abarbanel, the Knesset Chief Language Editor, for providing us with information of the transcription and language editing processes of the Parliament.
We are extremely grateful to Shira Wigderson and Lior Schwartz for their immense help preparing the Knesset databases. We would like to thank Shaul Shenhav, Guy Mor, Liran Harsgor and Alon Yakter for fruitful discussions. 
We thank the Idit PhD Fellowship Program at the University of Haifa for supporting the first author. 
This research was supported by the Ministry of Science \& Technology ,Israel under grant no.~3-17990. 

\section*{Limitations}
Like any corpus, the Knesset corpus is by no means representative, 
and in particular, is highly unbalanced. For example, around~81\% of the sentences were uttered  by men, and only about~19\% by women.
Hence, any conclusion resulting from processing the corpus is limited to the population of speakers it reflects and to the topics and the genres characterizing it. However, the sheer magnitude of the data probably guarantees the generality of observations gleaned from the dataset.

The automatic processing of the protocols relied on regular expressions and certain characteristic strings in the text in order to extract each speaker, their relevant texts, and other information. These methods were designed to fit the majority of document structures found in the dataset. However, as the protocols were manually typed by different people at different times, lacking a uniform format, each protocol is unique. Consequently, errors such as unrelated strings being misidentified as speaker names were unavoidable. Also, the process of automatically linking the speaker name as given in the protocol with the correct Knesset member is prone to errors. Although most of these automatically-established connections were manually reviewed and corrected, they may still contain errors. For instance, a speaker who is not a Knesset member may have been mistakenly linked to one of the Knesset members, or a certain Knesset member speaker may have not been correctly identified.

Furthermore, it is important to state that demographic meta-data such as gender and religion are only available for speakers who were identified as Knesset members, as we do not have this information for other speakers in the corpus. However, all of the features that were extracted directly from the texts, along with the morpho-syntactic annotations, are available for the entire data. 

In addition, the use of the wordfreq library in our study on diachronic changes may be limited, as it is trained on various sources that differ from the genre of the Knesset and may not accurately reflect the word frequency specific to our domain. Despite this constraint, our discovery of a change in lexical richness in Knesset proceedings over time aligns with findings on lexical density and richness change (albeit in a different direction) reported in other scholarly studies
on diachronic language change in scientific articles \citep{ZHOU2023101262} and literary works \citep{Stajner-12}. Notably, both studies investigated language change over longer periods than our Knesset corpus spans.

There are also potential effects of topical changes on lexical properties. Our ongoing research involves applying diachronic topic modeling to the Knesset corpus, which will help us examine whether and how the observed lexical trends are influenced by changes in topic diversity. However, prior research does not necessarily tie changes in lexical diversity and richness to topical changes
\citep{Stajner-12,ZHOU2023101262}.


\section*{Ethical Considerations}

The vast majority of meta-data associated with the members of the parliament (many of whom are alive) was retrieved from the official web portal of the Knesset Archives, made publicly accessible by the State of Israel. In the rare cases of missing details (e.g., year of birth, date of immigration), these properties were retrieved from the \href{https://oknesset.org/}{Open Knesset project} and relevant Wikipedia pages. 
All this information has been in the public domain.

Having said that, we acknowledge that our database of MPs, which aggregates demographic information on hundreds of past and present MPs, including their political orientation (which is derived directly from the orientation of the party they represent), can be abused. For example, with this database one can extract names of living MPs that subscribe to a specific political agenda; or group together MPs according to their religion.
Still, such capabilities do not require our database, since the same information could have been easily and directly obtained from public (and even official) resources such as Wikipedia or the Knesset archives. To ensure our dataset adheres to the highest standards of ethics and privacy protection, we consulted with our  Institutional Review Board (IRB) about releasing the MP table. They claimed that this was not an ethics issue but rather a privacy one, and referred us to the legal issue of maintaining and publishing \emph{data repositories}, which is regulated by Israel's Ministry of Justice. We submitted a query to the Ministry's \emph{Authority for Privacy Protection} and they responded that since we comply with the privacy policy of the sources from which the data were mined, we can freely release the database.

We employed three linguists (two women, one man, all native Hebrew speakers residing in Israel) to annotate a sample of 5,000 sentences for UD. 
The annotators were recruited directly (i.e., not via any crowd-sourcing platform) and were paid an hourly wage that is approximately 2.5 times the minimum wage in Israel. No human participants were required for this project.

No AI assistants were used in this project or in the write-up of this paper.

\bibliography{anthology,custom}

\appendix

\section{Evaluation of the extracted data}
\label{app:evaluation}
We first converted the raw data to valid documents.
In total, we managed to process~42,063 out of~44,027 files (95.5\%). Approximately~1600 documents were found to be duplicates. The remaining~350 documents were either empty, or the text was graphically presented. 
We believe the small number of unprocessed files is insignificant and does not impact the overall data integrity.

We developed a set of rules and regular expressions to extract structured data from the raw data of the Knesset protocols. To determine these rules, we used a small portion of the data which we sampled uniformly from the entire dataset, containing about~430 files (1\% of the documents). We manually analyzed the output of our extraction algorithm on these files in a trial-and-error process, making sure all of the outputs were valid, before running the algorithm on the entire data. To evaluate the validity of the data, we also manually checked the output on~20 randomly selected documents that were not part of the original subset, consisting of~30,636 sentences and~539 speakers. 

For each document, we checked the following details: whether an output was generated for this document, if the extracted date was accurate, and whether the first and last speaker were correctly identified, along with their first and last sentences. We also checked if all unique speakers mentioned in the document were identified as speakers in the output, and if the names that we listed for them were correct. In addition, we checked how many times some text was wrongly identified as a speaker name. 

We found that the output was generated for all the evaluation documents, and the date was correctly extracted in 100\% of them. The first and last speaker were also identified correctly in 100\% of the documents. The first sentence was correctly identified in all documents. The last sentence was correctly identified in all but one document. All the unique speakers were identified as speakers in the output and in most cases, their names were correctly spelled. 78 out of~539 speakers were accidentally written with a `\verb+<+' character at the beginning of the name, meaning the cleaning process has partly failed on those names. However, the correct names are still easily identifiable. Three speakers' names were not extracted properly and an empty string or a single character was extracted instead. A non-name text string was mistakenly identified as a speaker’s name 10 times in total, in~56 sentences, which are less than 0.183\% of the total number of sentences in the test subset. 

In addition, we evaluated the accuracy of the name matching between the speakers’ names as were extracted from the protocol and the MP names in our database. First, we extracted from 85\% of the documents a table consisting of~686,227 strings that were identified as speaker names.
We associated each such string with the ``clean'' form of the name, and with the name of the MP assigned to it, as it is listed in the MP database, if we found a match; otherwise, we listed the clean form of the name.
Two different types of error may occur: a wrong match may be found (either because the speaker is not a Knesset member or because we matched between two different Knesset members); or a match may not be found although the speaker is indeed a Knesset member. 
Out of the total set, 285,790 records were automatically detected as correct since a match was found and the assigned name was identical to the clean name. We manually analyzed the remaining~400K records. About~38K records (5.52\%) were found to be incorrect and were manually fixed. This reflects success rate of~94.48\% . 

Out of the~323,373 records that we found a match for, 18,192 were identified as wrong (false positives). Either the speaker was not a Knesset member or they were matched to a different MP. Most of these mistakes are caused by very similar names. 
Out of the~362,854 records that we did not find a match for, 19,686 were identified as wrong (false negatives): the speaker was indeed a Knesset member but was not identified as one. This was usually caused because the name, as was written in the protocol, is too different from the one listed in our database. 

It is important to note that all the identified mistakes were manually corrected and do not represent the number of mistakes currently in the corpus. However, we can expect similar results for the remaining 15\% 
of the documents that were not analyzed in this process. 








\section{Parsing}
\label{app:parsing}
\label{app:parsing-training}
Our training baseline was the UD treebank of \citet{zeldes-etal-2022-second}, significantly expanded with manually-annotated trees from diverse sources, as outlined in Table~\ref{tbl:treebank}.
These sources included the \emph{All Rights} entitlements awareness organization, the \emph{Davar} and \emph{Israel Hayom} news organizations, the \emph{GeekTime} Hebrew-language technology news site, and the Hebrew-language Wikipedia. Our unique contribution was the addition of manually-annotated Knesset data (excluding some 300 sentences due to technical constraints). 

\begin{table}[hbt]
\centering
\begin{tabular}{lrrr}
\textbf{Source} & \textbf{Sentences} & \textbf{Tokens} \\
All Rights & 12,639 & 167K \\
Davar & 13,699 & 223K \\
Geek Time & 4,923 & 96K \\
Israel Hayom & 1,466 & 16K \\
Wikipedia & 24,812 & 430K \\
\textbf{Knesset} & \textbf{4,691} & \textbf{69K} \\
\hline
Total & 62,230 & 1,001K \\
\end{tabular}
\caption{Breakdown of the treebanks used for parser training and evaluation, categorized by source.}
\label{tbl:treebank}
\end{table}

\label{app:parsing-evaluation}
Table~\ref{tbl:parsing-results}
lists the performance of the UD parser in terms of the metrics used by the \href{https://universaldependencies.org/conll18/evaluation.html}{CONLL-2018} evaluation script \citep{zeman-etal-2018-conll}.

\begin{table}[hbt]
\begin{tabular}{lrrr}
\textbf{Metric} & \textbf{Precision} & \textbf{Recall} & \textbf{F1 Score} \\
Tokens &99.88&99.85&99.86\\
Sentences &100.00&100.00&100.00\\
Words &99.54&99.57&99.55\\
UPOS &98.93&98.96&98.94\\
UFeats &97.38&97.41&97.39\\
UFeats+ &98.36&98.55&98.46\\
AllTags &96.59&96.61&96.60\\
Lemmas &98.79&98.82&98.81\\
UAS &97.78&97.81&97.79\\
LAS &97.16&97.18&97.17\\
CLAS &96.38&96.44&96.41\\
MLAS &91.81&91.86&91.83\\
BLEX &95.17&95.22&95.19\\
\end{tabular}
\caption{UD Parser performance.}
\label{tbl:parsing-results}
\end{table}

\begin{figure}[hbt]
\centering

\begin{subfigure}{0.5\textwidth}
    \includegraphics[width=\textwidth]{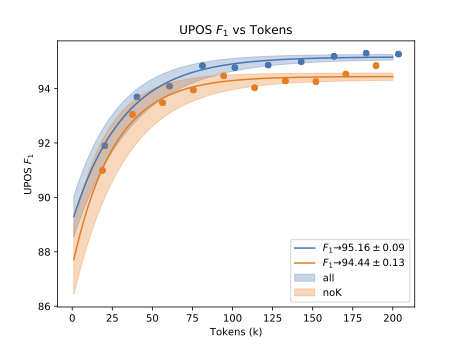}
    \caption{UPOS}
    \label{fig:iahlt-ud-curve-upos}
\end{subfigure}
\hspace{-0.5cm}
\begin{subfigure}{0.5\textwidth}
    \includegraphics[width=\textwidth]{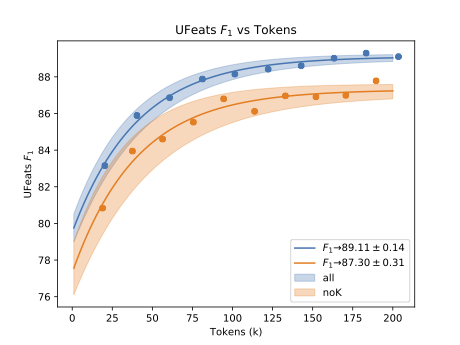}
    \caption{UFeats}
    \label{fig:iahlt-ud-curve-ufeats}
\end{subfigure} \\
\begin{subfigure}{0.5\textwidth}
    \includegraphics[width=\textwidth]{K-UAS.png}
    \caption{UAS}
    \label{fig:iahlt-ud-curve-uas}
\end{subfigure}
\hspace{-0.5cm}
\begin{subfigure}{0.5\textwidth}
    \includegraphics[width=\textwidth]{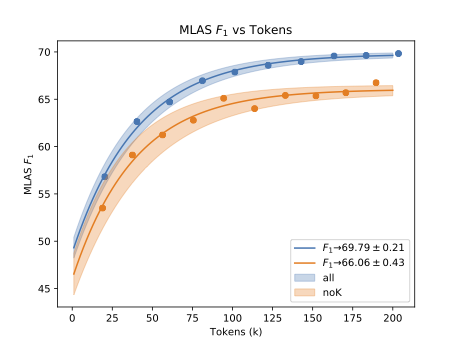}
    \caption{MLAS}
    \label{fig:iahlt-ud-curve-mlas}
\end{subfigure} 

\caption{Performance curves for models trained on subsets of a small
treebank, with and without the Knesset data, as evaluated on a sample
of Knesset sentences. }
\label{fig:iahlt-ud-curve}
\end{figure}

Figure~\ref{fig:iahlt-ud-curve}
shows the
performance curves for models trained on subsets of a small
treebank, with and without the Knesset data, as evaluated on a sample
of Knesset sentences. The inclusion of Knesset data consistently resulted in higher asymptotic performance, albeit with a slightly slower convergence rate. Error bars indicate variability from the curve fitting process.




\section{Dashboard Examples}
\label{app:dashboard}
\begin{figure}[hbt]
\centering
  \includegraphics[width=\textwidth]{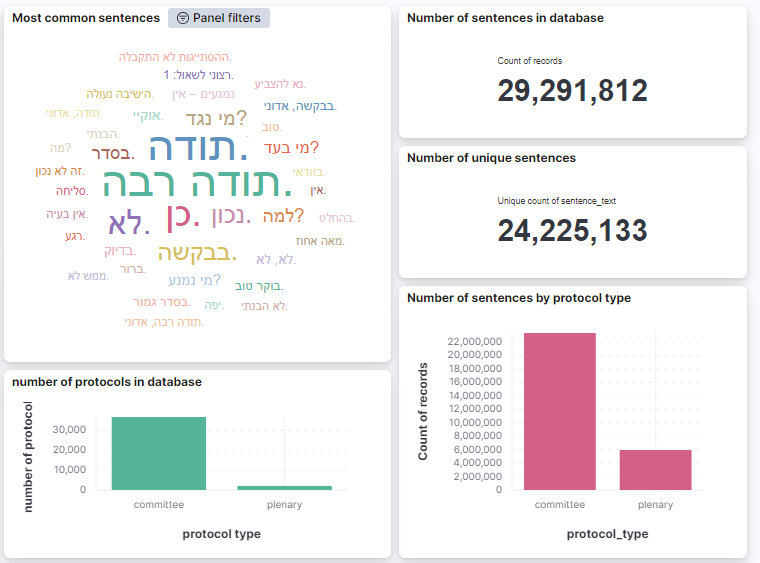}
  \caption{Dashboard example.}
  \label{fig:dashboard1}
\end{figure}

\begin{figure*}[hbt]
\centering
  \includegraphics[width=\textwidth]{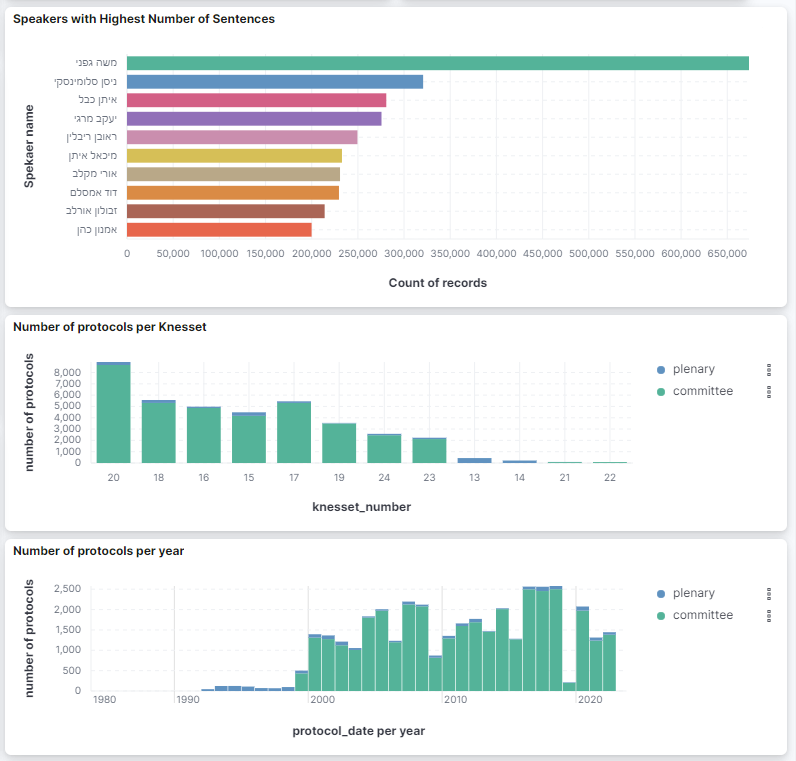}
  \caption{Dashboard example.}
  \label{fig:dashboard3}
\end{figure*}

Figures~\ref{fig:dashboard1} and~\ref{fig:dashboard3} show partial views of the dashboard
(Some figures in the dashboard examples are inconsistent with figures reported in the paper because the screen shots reflect an earlier version of the dataset.)

\section{Gendered Word Lists}
\label{app:word-lists}
\begin{figure*}[hbt]
\centering
  \includegraphics[width=0.7\textwidth]{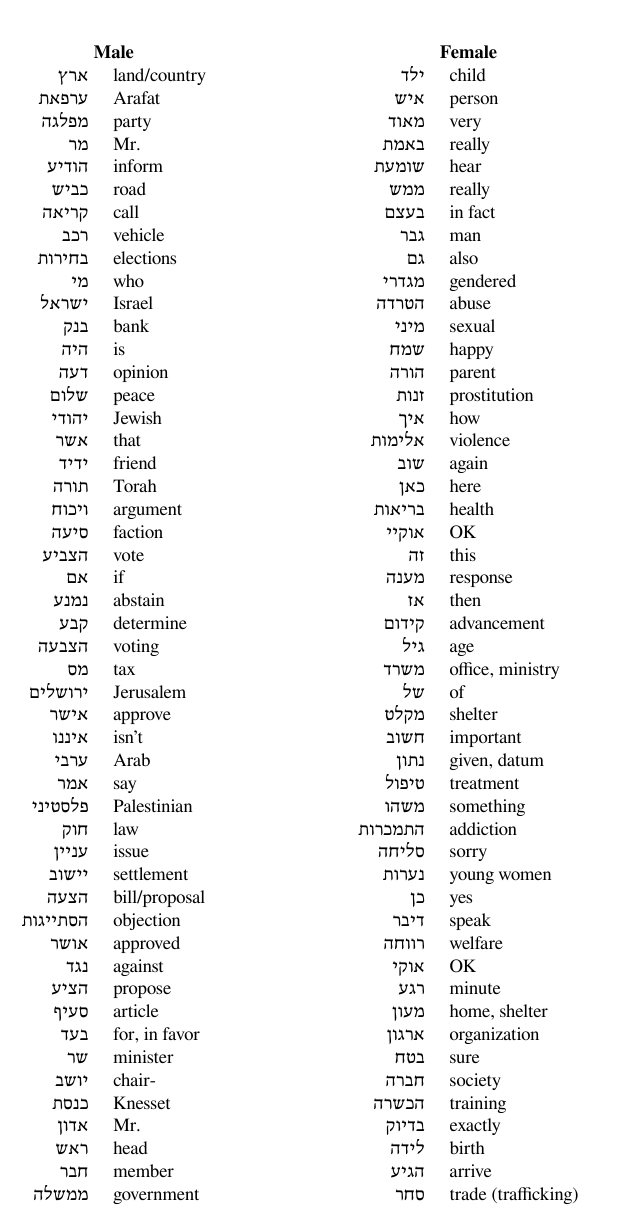}
  \caption{Hebrew word lists: top-50 most prominent words used by men and women.}
  \label{fig:word-lists}
\end{figure*}

Figure~\ref{fig:word-lists} depicts the top-50 most prominent words associated with men and women, respectively.

\end{document}